\relax
\documentclass{article}
\usepackage{times}
\usepackage{helvet}
\usepackage{courier}
\usepackage{algorithmic}
\usepackage{algorithm}
\usepackage{amsmath}
\usepackage{booktabs}
\usepackage{graphicx}
\usepackage{tikz}
\usepackage{pgfplots}
\pgfplotsset{compat=newest}
\usepackage{multirow}
\usepackage{natbib}
\begin{document}
%
\title{Exploiting Unlabeled Data for \\Neural Grammatical Error Detection}

\author{Zhuoran Liu$^\dagger$ and Yang Liu$^\ddagger$\\
$^\dagger$School of Software, Beihang University, Beijing\\
$^\ddagger$Department of Computer Science and Technology, Tsinghua University, Beijing\\
\texttt{liuzhuoran17@163.com, liuyang2011@tsinghua.edu.cn}}
\date{}
\maketitle
\begin{abstract}
Identifying and correcting grammatical errors in the text written by non-native writers has received increasing attention in recent years. Although a number of annotated corpora have been established to facilitate data-driven grammatical error detection and correction approaches, they are still limited in terms of quantity and coverage because human annotation is labor-intensive, time-consuming, and expensive. In this work, we propose to utilize unlabeled data to train neural network based grammatical error detection models. The basic idea is to cast error detection as a binary classification problem and derive positive and negative training examples from unlabeled data. We introduce an attention-based neural network to capture long-distance dependencies that influence the word being detected. Experiments show that the proposed approach significantly outperforms SVMs and convolutional networks with fixed-size context window. 
\end{abstract}

\section{Introduction}

\noindent Automatic grammatical error detection and correction for natural languages has attracted increasing attention, for a large number of non-native speakers are learning or using foreign languages. Take English as an example. There are a large number of English learners around the world who need instantaneous accurate feedback to help improve their writings (\cite{beam-decoder}). In the domain of scientific paper writing in which English is the main language, authors also need effective grammar checkers to help them in composing scientific articles (\cite{aesw-proposal}).

There have been several shared tasks addressing grammar errors in recent years. HOO-2011 (\cite{hoo-2011}), HOO-2012 (\cite{hoo-2012}), CoNLL-2013 (\cite{conll-2013}) and CoNLL-2014 (\cite{conll-2014}) shared tasks all aimed to correct grammar errors. The AESW shared task (\cite{aesw-2016}) aimed to identify sentence-level grammar error. These shared tasks helped advanced the research of grammatical error detection or correction.

Despite these advances, the scarcity of annotated data is still a major limitation on research of grammatical error detection and correction. Researchers need mass annotated data to train a grammar checker, but unfortunately for them, there are only a small amount of annotated corpora available in a limited number of domains. Most annotated corpora are in the domain of learner English, e.g. NUCLE (\cite{nucle}) and CLC (\cite{clc}), and others are from domains such as scientific papers, e.g. AESW dataset (\cite{aesw-proposal}). In order to train their system with enough data, researchers use multiple corpus instead of one (\cite{camb-conll-14}).

Data scarcity is partly due to difficulties in building an elaborately annotated corpus needed for training of a grammatical error correction system, as is described by the team that built NUS Corpus of Learner English (\cite{nucle}). In order to obtain a reliable annotation, they set up a guideline for annotators so that corrections are consistent. To ensure that these annotations are available, several annotators proposed their correction independently, and annotations most agreed upon are selected. Such annotating process is labour-intensive and time consuming, and the quality of the corpus are subject to human judgment and other factors such as budget. For example, their team is unable to perform double annotation for the main corpus due to budget constraints. They spent a long time (over half a year) to annotate only 1,414 essays.

Given these difficulties in building annotated corpus, we hope to utilize un-annotated error-free texts in unsupervised training of a grammatical error correction or grammatical error detection system. Previously, efforts have been made to explore how realistic grammatical errors could be counterfeited automatically from error-free texts and therefore obtain large amount of annotated data (\cite{ae4gec,generrate,GEMarkov,pseudo-error}). We therefore follow the idea of building a corpus by generating artificial error, since there are vast amount of un-annotated texts available and most of them are error-free. We explored two ways of artificial error generation, which is proved to be effective in our experiment.

Training a system to correct grammatical errors might be a more difficult task when there is no supervision, since there are numerous error types and our method to generate artificial errors might not be sophisticated enough to cover all of them. We thus focus on grammatical error detection instead of correction. It is natural to address this task as binary classification, in which we make prediction of whether a word is grammatically correct.

\section{Background}

\subsection{Problem Statement}
\label{prob-stmt}
The goal of word-level grammatical error detection is to identify grammar errors at the word level.
For example, given a sentence shown below, a grammatical error detection system is expected to correctly identify the erroneous word `{\em birds}' highlighted by underline:\\

\textit{An ugly \underline{birds} was observed by the man yesterday .}\\

The task of word-level grammatical error detection is formalized as such: given a sequence of token $X=(x_1, x_2, ..., x_n)$ as input, the error detector outputs its prediction $Y=(y_1, y_2, ..., y_n)$ where $y_i$ denotes the correctness of $x_i$ in terms of grammaticality.

We address this problem as a binary classification problem. In order to predict $y_t$ given the current word $x_t$ and the whole sentence $X=(x_1, x_2, ..., x_n)$, we need to find a function $g(\cdot)$ to calculate the conditional probability of each $y_t$ given $x_t$ and the whole input sequence $X$:
\begin{equation}
p( y_t \vert x_t) = g(x_t,X),
\end{equation}
where
\begin{align}
y_t = \left \lbrace
\begin{array}{r@{}l}
1 & \ \ \textrm{correct}\\
0 & \ \ \textrm{incorrect}
\end{array}.
\right.
\end{align}
Our aim is to build a suitable classification model for $g(\cdot)$.

\subsection{Support Vector Machines} \label{sec_svm}

A natural approach is to use Support Vector Machine (SVM) to perform classification (\cite{svm-1,svm-2}). SVM is trained given a training dataset in the form of $\{ (x_1,y_1), ... , (x_n,y_n) \}$, where $x_i$ represents a token with a set of selected linguistic features, and $y_i$ denotes the grammatical correctness of the token. It finds a maximum-margin hyperplane that separates correct words from incorrect ones.

The problem with this approach is that we need to manually design features in $x_i$. Since human are unable to tell precisely which features are relevant, human-designed features are inadequate in some aspects while being redundant in others. As a result, these designed features are unable to capture all regularities, which might hurt the performance of our error detector.

\subsection{Convolution Network with Fixed Window Size} \label{sec_cnn}

To circumvent the problem with feature engineering, a natural thought is to utilize the capability of neural networks in automatic feature extraction (\cite{cnn}). The simplest way is to take into consideration a fixed size window of words around the current word as its context by applying temporal convolution over the fixed size window.
In the example sentence given in Section 2.1, when considering the grammatical correctness of the word `\textit{was}' given a context window of size 3, the context window would be \textit{birds was observed}. The assumption that underlies this method is that only neighbouring words are grammatically related to the current word.

Here we formalize the method of neural network with fixed size window. Given a word $x_i$, its context $c_i$ is
\begin{equation}
c_i = (x_{i-w/2},...,x_{i},...,x_{i+w/2}).
\end{equation}

Let $f(\cdot)$ denote a temporal convolution operation with the input frame size equal to the dimension of $x_i$, the output frame size equal to $1$, and the kernel width equal to the size of fixed window. A score $s_i$ of the current word $x_i$ is calculated by $s_i = f(c_i)$ which represents grammatical features within the window. This score then goes through a sigmoid layer and yield the probability of $y_i$: $p(y_i | x_i, c_i) = \sigma (s_i)$.

The first problem with this method is that it is incapable of capturing long-distance dependency. With a fixed window size, the error detector is unable to take into consideration word contexts beyond the window size, while long-distance grammatical dependency is quite common a phenomena. For example, in order to determine whether `\textit{was}' is incorrect, we would need to take `\textit{yesterday}' into consideration, which requires a large size of context window.

Another problem with this approach is that all words within the context window is taken into consideration indiscriminately. In the example above, `\textit{was}' might not care about what was done to the birds when determining the verb tense, but `\textit{observed}' is given equal attention regardless of the fact that it has no influence on verb tense.

\begin{figure}[!h]
\centering
  \includegraphics[width=0.78\textwidth]{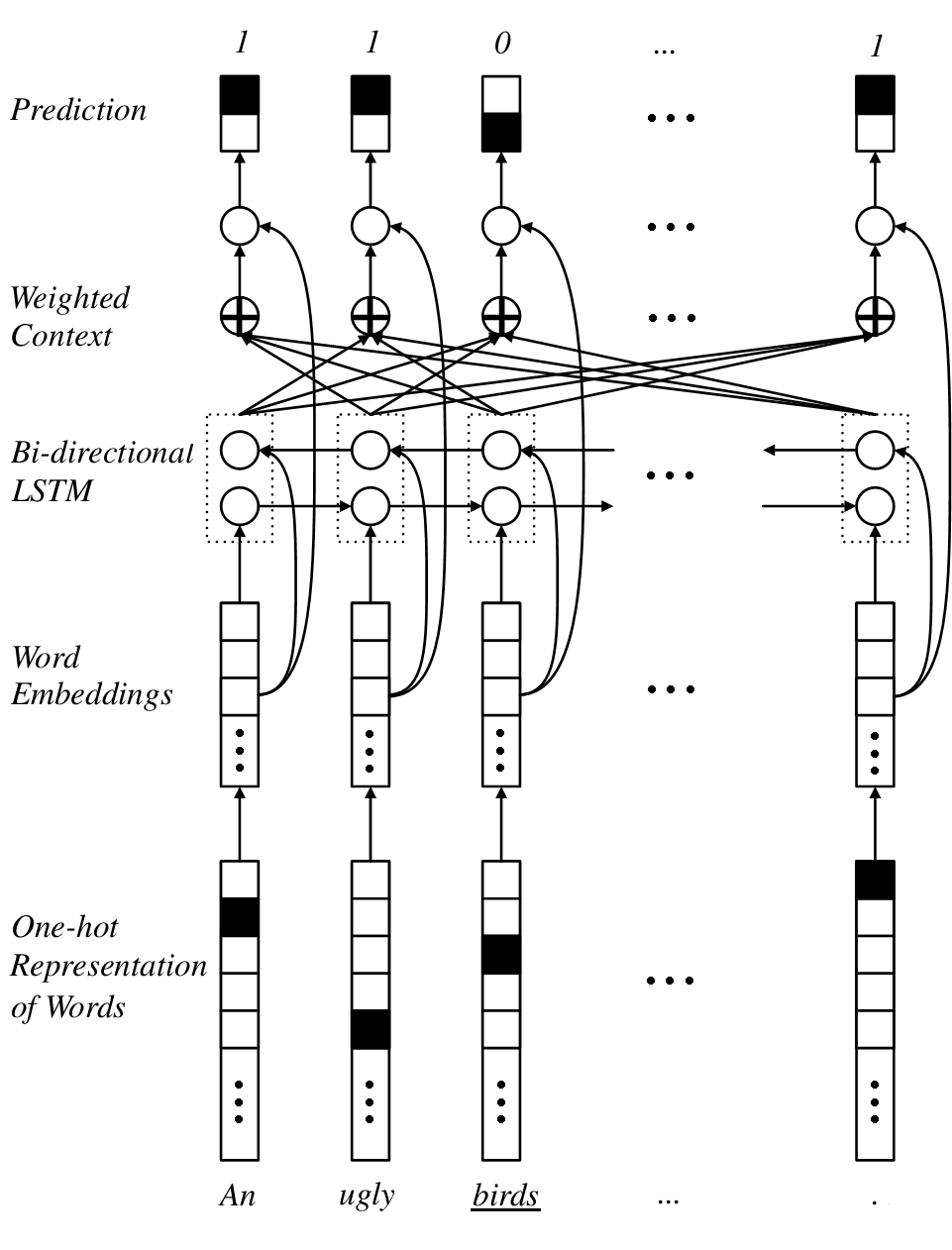}
  \caption{Model Architecture. The input of the network is a sentence \textit{An ugly birds was observed by the man yesterday .} in the form of one-hot representation. The representation is then converted into continuous word-embeddings and encoded by a bi-directional LSTM encoder. These encoded information is reweighted by an intra-attention mechanism at each time-step, on which the classifier judges the grammaticality of each word.}
  \label{fig:arch}
\end{figure}

\section{Approach}

In this section we describe our unsupervised approach to word-level grammatical error detection.

\subsection{Model Architecture}
Our intuition is to first encode the input sequence into a sequence of hidden states which contain relevant grammatical information, and then make prediction given a word and its context (see Figure.\ref{fig:arch}). Thus our model consists of two parts: an encoder that adopts a typical architecture of bi-directional LSTM network (\cite{lstm}), and a classifier that makes predictions based on hidden states of the encoder.

\subsubsection{Encoder}
The encoder takes as input a sentence $S$ of length $n$, represented by a sequence of vector $X=(x_1,x_2,...,x_n)$. In an LSTM recurrent neural network, the input $X$ is processed through time and produces a series of memory states $(c_1,c_2,...,c_n)$ and hidden states $(h_1,h_2,...,h_n)$. In order to counterbalance the impact of time on hidden states we process the input $X$ two times, forward and backward, to fully encode the information classifier needs.

The forward LSTM updates its memory state $\overrightarrow{c_i}$ and hidden state $\overrightarrow{h_i}$ at each time step $t$:
\begin{equation}
[\overrightarrow{h_t}; \overrightarrow{c_t}] = \overrightarrow{LSTM} ([\overrightarrow{h_{t-1}};\overrightarrow{c_{t-1}}]).
\end{equation}

Similarly, memory state $\overleftarrow{c_i}$ and hidden state $\overleftarrow{h_i}$ is updated by backward LSTM at time step $t$:
\begin{equation}
[\overleftarrow{h_t}; \overleftarrow{c_t}] = \overleftarrow{LSTM} ([\overleftarrow{h_{t+1}};\overleftarrow{c_{t+1}}]).
\end{equation}

The encoder outputs a hidden tape $\widetilde{h} = (\widetilde{h_1},\widetilde{h_2},...,\widetilde{h_n})$, where
\begin{equation}
\widetilde{h_t} = \left[
\begin{matrix}
\overrightarrow{h_t}\\
\overleftarrow{h_t}
\end{matrix}
\right].
\end{equation}

\subsubsection{Classifier with Intra-attention}

To predict whether the word at time step $t$ is grammatically problematic, the classifier computes a score given the current word $x_t$ and its context $a_t$. This score $s_t$ then goes through a sigmoid layer and makes a binary prediction, with 1 denoting grammatically correct and 0 denoting incorrect. Note that the Classifier does not hold its own state as a decoder does in a traditional encoder-decoder architecture.

To address the problem of long-distance dependency, we incorporate an intra-sentence attention mechanism (\cite{rnnsearch}) in our classifier, where all hidden states of encoder is taken into consideration and the attention of classifier on all position of the sentence is dynamically adapted. To describe formally, we compute the context $a_t$ around the word $x_t$ as an attention-weighted sum of $\{\widetilde{h_1},\widetilde{h_2},...,\widetilde{h_n}\}$:
\begin{equation}
a_t = \sum_i \alpha_{t,i} \cdot \widetilde{h_i},
\end{equation}
where
\begin{equation}
\alpha_{t,i} = \frac{\exp(E_{t,i})}{\sum_j \exp(E_{t,j})},
\end{equation}
\begin{equation}
E_{t,i} = \widetilde{h_t}^\top \cdot \widetilde{h_i}.
\end{equation}

Vector $a_t$ represents the grammatical and semantic context at position $t$. A word is considered to be grammatically erroneous if the word $x_t$ does not fit into the current context, i.e. it is incompatible to place $x_t$ at position $t$ given the context $a_t$. The score $s_t$ is computed as follows:
\begin{equation}
s_t = x_t^\top \cdot W \cdot a_t + b.
\end{equation}
where $b$ is the bias.

Then the probability can be calculated as
\begin{equation}
p( y \vert x_t , \{ x_1, x_2, ... , x_n \} ) = \sigma ( s_t ).
\end{equation}

By incorporating intra-attention mechanism, we provide a latent structure for the model to learn grammatical relations between words. This makes a lot of sense because the grammaticality of a word is dependent more on the words that have strong grammatical relation with it, while others are negligible when making prediction. For example in Figure.\ref{fig:arch}, when the model tries to determine whether `\textit{birds}' is correct in terms of noun number, it will pay a strong attention to `\textit{An}'.

\subsection{Noise Generation}

Traditionally, a large set of $\{ \langle X^{(n)} , Y^{(n)} \rangle \}^N_{n=1}$ is needed to effectively train such a grammatical error detector. However in an unsupervised approach, only $\{ X^{(n)} \}^N_{n=1}$ is given. The key issue is how the corresponding $\{ Y^{(n)} \}^N_{n=1}$ could be obtained.

We adopted the idea of using artificial error for training. It is crucial to find a suitable algorithm for the error generator to produce realistic grammatical error, since the performance of the model relies heavily on the paradigm it saw during training. Since our task is to detect grammatical error on word level, we only consider substitution errors. We compare two ways of substituting the original word for an erroneous one.
\subsubsection{Uniform random substitution}
The simplest way is to substitute a word in a random position with a random word from the vocabulary. The problem with this approach is that some artificial errors generated in this way is apparently irrelevant. For example, it could substitute a word from the sentence\\

\textit{An ugly bird was observed by the man yesterday .}\\

\noindent to generate such a sentence as\\

\textit{An ugly bird was \textbf{vocabulary} by the man yesterday .}\\

One potential problem is that it might be too easy for our classifier to discriminate such erroneous words from the correct ones.

\begin{algorithm}[!h]
\caption{Build Substitution Set}
\label{alg1}
\begin{algorithmic}
\STATE PoS-tag the input text
\STATE Build a dictionary $D$ of (token, PoS-tag)
\FORALL{$(token,pos)$ in $D$}
\IF{$pos$ in \{CC\} \OR \{DT, PDT\} \OR \{PRP, PRP\$\} \OR \{IN, TO, RP\} \OR \{WDT, WP, WP\$, WRB\}}
\STATE Add $token$ to the corresponding substitution set $c_i$
\ELSIF{$pos$ in \{NN, NNP, NNPS, NNS\} \OR \{VB, VBD, VBG, VBN, VBP, VBZ\}}
\STATE $lemma \leftarrow$ Lemmatise $token$
\STATE Add $token$ to the corresponding substitution set $c_i$
\ELSIF{$pos$ in \{JJ, JJR, JJS\} \OR \{RB, RBR, RBS\}}
\STATE $stem \leftarrow$ Stem $token$
\STATE Add $token$ to the corresponding substitution set $c_i$
\ENDIF
\ENDFOR
\end{algorithmic}
\end{algorithm}

\begin{algorithm}[!h]
\caption{Error generation}
\label{alg2}
\begin{algorithmic}
\FORALL{sentence $S$ in training text}
\STATE Get word $w$ at random position of $S$
\STATE $w' \leftarrow w$
\STATE Search for substitution set $c_i$ that contains $w$
\IF {such $c_i$ does not exist \OR $c_i$ contains only 1 element}
\WHILE{$w' == w$}
\STATE $w' \leftarrow$ Select a random word from dictionary $D$
\ENDWHILE
\ELSE
\WHILE{$w' == w$}
\STATE $w' \leftarrow$ Select a random word from $c_i$
\ENDWHILE
\ENDIF
\STATE Replace $w$ in $S$ with $w'$
\ENDFOR
\end{algorithmic}
\end{algorithm}

\begin{table*}[!h]
\centering
\begin{tabular}{c|c|l|p{0.4\columnwidth}}
\toprule
  \textbf{POS tag} & \textbf{Original Word} & \textbf{Noise} & \textbf{Example}\\
\midrule
\midrule
  \multirow{2}{*}{VB} & \multirow{2}{*}{built} & \multirow{2}{*}{build, builds, building, ...} & \textit{Workers built the park centuries ago .}\\
  & & & \textit{Workers \textbf{build} the park centuries ago .} \\
\midrule
  \multirow{2}{*}{NN} & \multirow{2}{*}{eggs} & \multirow{2}{*}{egg} & \textit{All eggs were put into the same basket .}\\
  & & & \textit{All \textbf{egg} were put into the same basket .} \\
\midrule
  \multirow{2}{*}{DT} & \multirow{2}{*}{an} & \multirow{2}{*}{a, this, these, ...} & \textit{There is an apple on the table .} \\
  & & & \textit{There is \textbf{a} apple on the table .} \\
\midrule
  \multirow{2}{*}{RB} & \multirow{2}{*}{suitably} & \multirow{2}{*}{suitable} & \textit{Candidates must be suitably qualified students.} \\
  & & & \textit{Candidates must be \textbf{suitable} qualified students.}\\
\midrule
  \multirow{2}{*}{IN} & \multirow{2}{*}{of} & \multirow{2}{*}{in, by, for, at, ...} & \textit{This book consists of 12 chapters .} \\
  & & & \textit{This book consists \textbf{by} 12 chapters .} \\
\bottomrule
\end{tabular}
\caption{Examples of substitution with linguistic knowledge.}
\label{tab:error-ling}
\end{table*}

\subsubsection{Substitution with linguistic knowledge}
We carefully examined a number of erroneous paradigms and found some characteristics common to all grammatical errors, regardless of the terminology and commonly seen patterns of the domain. To briefly summarize it, errors usually appears when a correct word is substituted by another word, which comes from a finite set of words and that are linguistically related to it, either because they possess the same lemma or the same part-of-speech tag. There is an inexhaustible list of how linguistic knowledge works in substitution. Here we only present several examples in Table.\ref{tab:error-ling}.

Combining these two method of error generation, we are able to generate 16 types grammatical errors out of 28 specified by CoNLL-2014 Shared Task (\cite{conll-2014}). \footnote{The error types can be generated are: Vt, Vm, Vform, SVA, ArtOrDet, Nn, Npos, Pform, Pref, Prep, Wci, Wform, Spar, Trans, Mec, Others. Most of remaining error types we are unable to generate are either semantic errors (Smod, Rloc-, UM), or style problems (Wa, Wtone, Cit), or sentence level problems (Srun, Sfrag, WOinc, WOadv).} Details are described by Algorithm \ref{alg1}, which formalizes the construction of substitution set, and Algorithm \ref{alg2}, which formalizes the process of error-generation with linguistic knowledge.

\section{Experiments}
\subsection{Settings}

\subsubsection{Data}
We used data mainly from three sources (Table \ref{tab:data}):
\begin{itemize}
\item ACL Anthology\footnote{http://www.aclweb.org/anthology/} (ACL): training set.
\item AESW Shared Task Dataset (AESW) (\cite{aesw-2016}) : development and test sets.
\item CCL Anthology\footnote{http://www.cips-cl.org/anthology} (CCL): development and test sets.
\end{itemize}

\begin{table}[!h]
  \centering
  \begin{tabular}{cc|rrrr}
    \toprule
    & \textbf{Set} & \textbf{\# Token} & \textbf{Pct.} & \textbf{Vocab.} & \textbf{\# Sent.}\\

  	\midrule
  	\midrule
  	\multirow{2}{*}{AESW} & dev & 24.4K & 6.0\% & 4.1K & 1.0K
\\
						 & test & 24.7K & 5.9\% & 4.1K & 1.0K
\\
  	\midrule
  	\multirow{2}{*}{CCL} & dev & 2.6K & 5.5\% & 892 & 125
\\
						 & test & 2.8K & 5.2\% & 934 & 126
\\
  	\midrule
  	ACL & train & 60.4M & 4.9\% & 166.5K & 2.9M
\\
  	\bottomrule
  \end{tabular}
  \caption {Statistics of datasets used in the experiment. ``Pct." stands for percentage of tokens that are marked incorrect.}
  \label{tab:data}
\end{table}

For training set, we used sentences from papers that appear in ACL Anthology. We crawled all papers up to year 2015, and then select sentences that end with a period, with a length of longer than 5 but no longer than 50, which may contain several clauses separated by commas, colons or semicolons. Formulae and references are excluded, numbers are substituted with a special $\langle num\rangle$ token, and parentheses are removed together with the contents in between. We limit the vocabulary to tokens with at least a word-frequency of 2 to eliminate most spelling errors, and replaced all OOVs with a special $\langle unk\rangle$ token.

To corroborate that the model trained by us actually works with realistic grammatical error, we used two human annotated dataset as our development and test set.

The first one is the test set of AESW 2016 Shared Task, but we only used a portion of the erroneous sentences from paragraphs with the attribute of ``domain=Computer Science"; we converted the data format by preserving all words between `$\langle del\rangle\langle /del \rangle$' and marking them as incorrect, while removing those between `$\langle ins\rangle\langle /ins\rangle$'.

For example, if the original annotated sentence is\\

\textit{More discussions $\langle del \rangle$about$\langle /del\rangle\langle ins \rangle$on$\langle /ins\rangle$ these issues will be provided in the remainder of the monograph.}\\

\noindent We convert it into the form of:\\

\textit{More discussions \textbf{about} these issues will be provided in the remainder of the monograph.}\\

The second human annotated dataset is some erroneous sentences from papers in CCL Anthology annotated by us, which contains grammatical errors since most of them are written by Chinese.

\subsubsection{Baselines}
To the best of our knowledge, word-level grammatical error detection task has never been researched before. Thus we use the two methods described in \ref{sec_svm} and \ref{sec_cnn}. We compared our method to two baselines, both of which are trained on the ACL training set.

\begin{itemize}
\item Support Vector Machine (SVM).
\item Convolutional Network (Conv).
\end{itemize}

SVM takes into consideration the context in a fixed window of size 5 around the current word, and gives prediction of whether the current word is grammatically correct in the sentence. We first trained an $n$-gram model with KenLM (\cite{kenlm}) on the whole training set without artificial errors, with $n$ up to 3. We then use $n$-gram scores as the input features into the SVM. In our experiment we used the open-source tool LibLinear (\cite{liblinear}).

In Conv, we use word-embeddings pre-trained using word2vec model in gensim (\cite{gensim-word2vec}), the dimensionality of which we set to 50 empirically. A temporal convolution is performed over a window of fixed size 3. The kernel width is set to equal to the size of fixed window. This model is implemented using Torch7. \footnote{http://torch.ch/}

\subsubsection{BiLSTM with Intra-attention}
Our model is implemented using Tensorflow.\footnote{https://www.tensorflow.org/} We used cross-entropy as our loss function to optimize. We perform gradient clipping by global norm (\cite{gradient-clipping}) with the function provided in Tensorflow. The dimention of word-embedding and hidden state are set to 150, as a trade-off between performance and training time. The word-embedding matrix is initialized with random uniform distribution within range of $\pm 0.05$.

\begin{table}[!h]
  \centering
  \begin{tabular}{c|c|rrr}
    \toprule
    \textbf{Method} & \textbf{Noise} & \textbf{Precision} & \textbf{Recall} & $\mathbf{F_{0.5}}$ \\

  	\midrule
  	\midrule
  	\multirow{2}{*}{SVM} & uni. & 13.53 & 6.27 & 10.99\\
						 & ling. & 12.51 & 7.15 & 10.88\\
  	\midrule
  	\multirow{2}{*}{Conv} & uni. & 6.25 & 50.10 & 7.57\\
						 & ling. & 18.13 & 4.46 & 11.24\\
  	\midrule
  	\multirow{2}{*}{BiLSTM} & uni. & 17.16 & 5.39 & 11.95\\
						 & ling. & 18.71 & 7.48 & \textbf{14.40}\\
  	\bottomrule
  \end{tabular}
  \caption {Performance on the AESW test set measured by {$F_{0.5} (\%)$}.}
  \label{tab:aesw}
\end{table}

\begin{table}[!h]
  \centering
  \begin{tabular}{c|c|rrr}
    \toprule
    \textbf{Method} & \textbf{Noise} & \textbf{Precision} & \textbf{Recall} & $\mathbf{F_{0.5}}$ \\

  	\midrule
  	\midrule
  	\multirow{2}{*}{SVM} & uni. & 7.40 & 1.34 & 3.89\\
						 & ling. & 6.25 & 1.34 & 3.61\\
  	\midrule
  	\multirow{2}{*}{Conv} & uni. & 5.66 & 57.43 & 6.91\\
						 & ling. & 6.66 & 0.67 & 2.4\\
  	\midrule
  	\multirow{2}{*}{BiLSTM} & uni. & 16.00 & 2.68 & 8.03\\
						 & ling. & 21.05 & 8.05 & \textbf{15.91}\\
  	\bottomrule
  \end{tabular}
  \caption {Performance on the CCL test set measured by {$F_{0.5} (\%)$}.}
  \label{tab:ccl}
\end{table}

\subsection{Results and Discussion}
Tables \ref{tab:aesw} and \ref{tab:ccl} presents the results of experiments of two baselines (SVM and T-Conv) and our model (BiLSTM), using uniform random errors (uni.) or errors counterfeited with linguistic knowledge (ling.). From the two tables we can see that our model outperformed the two baselines on both human-annotated datasets (AESW and CCL). The $F_{0.5}$ score might seem low, but they are actually good results since these models are trained without supervision.

\subsubsection{Effect of Error Types}
If we focus on the task of detecting a limited number of error types (Verb form, Noun Number, Preposition misuse, Article misuse), the model would show better performance on CCL test set, but weaker or only comparable to AESW test set. This is probably because in the annotating phase of CCL test set, we focused heavily on these common types of errors and some other types of errors are neglected. The results are shown in Tables \ref{tab:aesw-cmp} and \ref{tab:ccl-cmp}.

\begin{table}[!h]
  \centering
  \begin{tabular}{c|c|rrr}
    \toprule
    \textbf{Error Types} & \textbf{Attention} & \textbf{Precision} & \textbf{Recall} & $\mathbf{F_{0.5}}$ \\

  	\midrule
  	\midrule
  	\multirow{2}{*}{All} & w/o & 14.84 & 6.61 & 11.88\\
						 & w/ & 18.71 & 7.48 & 14.40\\
  	\midrule
  	\multirow{2}{*}{Limited} & w/o & 19.87 & 4.25 & 11.45\\
						 & w/ & 18.48 & 6.27 & 13.31\\
  	\bottomrule
  \end{tabular}
  \caption {Comparison on the AESW test set.}
  \label{tab:aesw-cmp}
\end{table}

\begin{table}[!h]
  \centering
  \begin{tabular}{c|c|rrr}
    \toprule
    \textbf{Error Types} & \textbf{Attention} & \textbf{Precision} & \textbf{Recall} & $\mathbf{F_{0.5}}$ \\

  	\midrule
  	\midrule
  	\multirow{2}{*}{All} & w/o & 23.40 & 7.38 & 16.32\\
						 & w/ & 21.05 & 8.05 & 15.91\\
  	\midrule
  	\multirow{2}{*}{Limited} & w/o & 26.00 & 8.72 & 18.62\\
						 & w/ & 27.90 & 16.10 & 24.34\\
  	\bottomrule
  \end{tabular}
  \caption {Comparison on the CCL test set.}
  \label{tab:ccl-cmp}
\end{table}

\subsubsection{Effect of Attention}

To verify our intra-attention help improve model performance, we removed the attention and performed the same experiment. Comparison of models with and without attention is shown in Tables \ref{tab:aesw-cmp} and \ref{tab:ccl-cmp}.

Though the intra-attention mechanism worked, but in some cases it may fail. We believe that a more sophisticated way of error generation is needed, because currently only those positions where substitution happens have a chance to be labeled incorrect (``on-site-error" paradigms). But for the model to learn grammatical relations by attention mechanism, we need massive paradigms where substitution cause another position to be labeled incorrect (``off-site-error" paradigms). For example, in the sentence\\

\textit{An ugly bird was observed by the man yesterday .}\\

\noindent If we substitute \textit{ugly} with \textit{beautiful}, our system will automatically annotate \textit{beautiful} as incorrect (on-site-error)\\

\textit{An \textbf{\underline{beautiful}} bird was observed by the man yesterday .}\\

\noindent but our model will never know why it is incorrect. What we need instead are off-site-errors:\\

\textit{\underline{An} \textbf{beautiful} bird was observed by the man yesterday .}\\

\noindent so that the model knows \textit{beautiful} is ok with `\textit{A}' but not with `\textit{An}'. Unfortunately our method does not provide such a mechanism to massively produce ``off-site-error" paradigms, therefore our model have to rely on very few coincidentally generated ``off-site-error" paradigms which are too sparse.

Our current method also introduced some substitutions which should not be counted as errors. For example, if \textit{yesterday} is substituted by \textit{today}:\\

\textit{An ugly bird was observed by the man \textbf{\underline{today}} .}\\

\noindent `\textit{today}' is annotated as incorrect under our method, while it does not actually constitute grammatical error, which therefore hurt model performance to some degree.

\subsubsection{Examples}

Our model is found to perform well in some cases, while failing to identify others (Table \ref{tab:examples})\footnote{The table contains only a partial list of error type our model detected. Since we only detect errors without inferring their types, we are unable to provide the full list of error types our model is able to detect.}. It works well with collocations, as seen in Ex.1. It also works well with morphological problems, as Ex.2 shows. However it is incapable of detecting errors of  genre as in Ex.3. In Ex.4, it mistook as incorrect words out of domain of the training data. It is apparent in Ex.5 that our model is aware of the missing \textit{are} between \textit{polarities} and \textit{opposite}, but it reports error at a different position. There are other error types our model did not handle well with, such as redundant determiners as in Ex.6.

\begin{table*}[!h]
\centering
\begin{tabular}{c|p{0.8\columnwidth}}
\toprule
  \textbf{Error Type} & \textbf{Example}\\
\midrule
  Collocation & (Ex.1) \textit{In \underline{\textbf{additions}} , we present an in-depth analysis that provides valuable insight into the characteristics of alternative solutions.} \\
\midrule
  Morphology & (Ex.2) \textit{In our work, lexical level features include the two entities, their NER tags, and the \underline{\textbf{neighbor}} tokens of these two entities.} \\
\midrule
  Genre & (Ex.3) \textit{For the purpose of this study, we focus on one of our live \underline{\textbf{broadcast}}, Premier Wen Talks Online with Citizens on \textbf{Feb} 28, 2009.} \\
\midrule
  Domain & (Ex.4) \textit{Note that , the communication cost of the PAROS layer is constant and dependent on the network size and the \underline{gossiping} period.} \\
\midrule
  Wrong Position & (Ex.5) \textit{In summary, filtering the sentences whose \underline{polarities} \textbf{opposite} to the overall orientation is significant for constructing a high quality training set.} \\
\midrule
  Other & (Ex.6) \textit{Compared with other methods, the \textbf{our} heterogeneous graph method improves the results significantly.} \\
\bottomrule
\end{tabular}
\caption{Example model predictions. Incorrect words are in bold face, and errors detected by our model are highlighted by underlines.}
\label{tab:examples}
\end{table*}

\section{Related Work}

\subsection{Grammatical Error Detection and Correction}
Several shared tasks on grammatical error detection or correction have been carried out in recent years, including HOO-2011 (\cite{hoo-2011}), HOO-2012 (\cite{hoo-2012}) , CoNLL-2013 (\cite{conll-2013}), and CoNLL-2014 (\cite{conll-2014}). These four shared tasks all focused on grammatical error correction of English written by non-native speakers. The AESW shared task (\cite{aesw-2016}) proposed to evaluate scientific writing automatically based on sentence-level error identification.

Different from these shared tasks, we focus on word-level grammatical error detection, which is a pilot step towards unsupervised approach to error correction.

To address the issue of grammatical error, researchers explored and utilized various methods. For example classification method was used by the top ranking team (\cite{ui-conll-13}) in CoNLL-2013 shared task. The top ranking team (\cite{camb-conll-14}) of CoNLL-2014 shared task incorporated in their system a Statistical Machine Translation (SMT) component, which translates erroneous English into correct English. With the development of Neural Machine Translation (NMT) and attention mechanism (\cite{rnnsearch}), the top team (\cite{sentence-level-ge}) of AESW shared task adopted the NMT approach to grammatical error correction.

In comparison, we adopted a typical architecture of bi-directional LSTM (\cite{lstm}) on the encoder side (\cite{cho-encdec}), but replace the decoder with a classifier. Since error types are not given in unsupervised training, our classifier does not infer error types but only make binary prediction.

\subsection{Error Generation}
To obtain enough training data, various approaches have been employed to generate artificial error. Markov logic network was used for statistical grammar error simulation (\cite{GEMarkov}). An automatic tool for error generation was developed (\cite{generrate}), which take as input a corpus and error generation rules. Error inflation was used in UI system (\cite{ui-hoo-12}) in HOO-2012 shared task, and similar method was performed on Japanese (\cite{pseudo-error}). To enlarge the size of training set, artificial errors were injected into the corpus by Yuan and Felice in CoNLL-2013 shared task (\cite{constrained-conll-13}). Later they further researched the probabilistic manner of artificial error generation with linguistic information (\cite{ae4gec}).

Different from (\cite{generrate}), we build substitution set automatically from un-annotated corpus based on POS tag or lemma, while their tools require a set of rules to work. To compare with (\cite{ui-hoo-12,constrained-conll-13,ae4gec}), their methods of error generation are based on annotated corpus, while we used only un-annotated error-free texts without any supervision.

\subsection{RNNs and LSTM Units}
Recurrent Neural Network (RNN) with Long Short-term Memory (LSTM) or Gated Recurrent Unit (GRU) has shown a mighty capability to encode informations over long sequences (\cite{cho-encdec}). The attention mechanism has enabled a Bi-directional RNN with GRU to achieve even better performance in machine translation (\cite{rnnsearch}) by allowing the decoder to explicitly make use of the memory of encoder. Upon the emergence of attention mechanisms, it has been applied to many NLP topics other than machine translation. Grammatical error correction is of no exception. Schmaltz et al. used a uni-directional LSTM network with attention mechanism and achieved the best performance in the AESW shared task (\cite{sentence-level-ge}).

Different from them, we do not need to generate target sentence because we are not doing correction. Therefore we replaced the decoder with a binary classifier, which take into consideration the information from BiLSTM encoder.

\section{Conclusion and Future Work}
In our work, we have explored unsupervised word-level grammatical error detection using only un-annotated corpus as training data. We showed that it is a viable way for machines to learn grammatical relations and to predict grammatical errors. This inspires us to further extend unsupervised approach to grammatical error correction. In the future, we plan to investigate novel methods for generating artificial errors to enable our model to learn better intra-sentence attention.

\nocite{nucle,confusion,memn2n}

\selectfont
\bibliography{aaai17}
\bibliographystyle{named}

\end{document}